\documentclass{article}
\usepackage{spconf,amsmath,graphicx}

\usepackage{enumitem}
\setlist{nosep, leftmargin=14pt}

\usepackage{mwe} 
\usepackage{xcolor}
\usepackage{amssymb}
\usepackage{multirow}
\usepackage[normalem]{ulem}
\usepackage{hyperref}       

\title{Post-hoc Overall Survival Time Prediction from Brain MRI}

\name{Renato Hermoza$^{\dagger}$ \qquad Gabriel Maicas$^{\dagger}$ \qquad  Jacinto C. Nascimento$^{\ddagger}$ \qquad Gustavo Carneiro$^{\dagger}$}
\address{$^{\dagger}$Australian Institute for Machine Learning, The University of  Adelaide \\ $^{\ddagger}$Institute for Systems and Robotics, Instituto Superior Tecnico, Portugal}

\begin{document}
%
\maketitle
\begin{abstract}

Overall survival (OS) time prediction is one of the most common estimates of the prognosis of gliomas and is used to design an appropriate treatment planning.
State-of-the-art (SOTA) methods for OS time prediction follow a pre-hoc approach that require computing the segmentation map of the glioma tumor sub-regions (necrotic, edema tumor, enhancing tumor) for estimating OS time.
However, the training of the segmentation methods require ground truth segmentation labels which are tedious and expensive to obtain.
Given that most of the large-scale data sets available from hospitals are unlikely to contain such precise segmentation, those SOTA methods have limited applicability.
In this paper, we introduce a new post-hoc method for OS time prediction that does not require segmentation map annotation for training.
Our model uses medical image and patient demographics (represented by age) as inputs to estimate the OS time and to estimate a saliency map that localizes the tumor as a way to explain the OS time prediction in a post-hoc manner.
It is worth emphasizing that although our model can localize tumors, it uses only the ground truth OS time as training signal, i.e., no segmentation labels are needed.
We evaluate our post-hoc method on the Multimodal Brain Tumor Segmentation Challenge (BraTS) 2019 data set and show that it achieves competitive results compared to pre-hoc methods with the advantage of not requiring segmentation labels for training.
We make our code available at \href{https://github.com/renato145/posthocOS}{https://github.com/renato145/posthocOS}.

\end{abstract}
\section{Introduction}
\label{sec:intro}

Gliomas are the most common brain tumors in adults, where the more severe form of the disease, high-grade gliomas, can cause a short life expectancy with a median of two years or less~\cite{menze_multimodal_2015}. 
Treatment planning for gliomas aims to increase the overall survival (OS) time. Thus, accurate estimate of the OS time is important because it affects treatment decision and patient prognosis.

Recent methods~\cite{agravat2019brain,wang2019automatic,feng2019brain,wang20193d,zhou2020m2net} employed for OS time prediction follow a pre-hoc pipeline~\cite{maicas2019pre} that consists of two stages. 
In the first stage, tumors are segmented into the following tissues: necrotic, edema tumor or enhancing tumor. 
Next, hand-designed radiomic features are extracted to train a OS time prediction model. 
However, the dependence on manually~\cite{zhou2020m2net} or automatically obtained tumor segmentation maps~\cite{agravat2019brain,wang2019automatic,feng2019brain,wang20193d} represents an important weakness of pre-hoc approaches because not only are these segmentation maps tedious and expensive to be acquired, but they can also contain noise due to annotation disagreements between experts~\cite{bakas_identifying_2019}. 
Furthermore,
the majority of data sets available from hospitals and clinics do not contain such segmentation maps. 
Hence, requiring segmentation labels limits the applicability of pre-hoc methods that predict patient OS time.

In this paper, we propose a new post-hoc method~\cite{maicas2019pre} for OS time prediction using as inputs medical image and patient demographics (represented by age) that is trained with only the OS time label, without requiring segmentation labels.
Our model consists of two parts that run in parallel. One part automatically estimates the OS time, while the other part highlights the relevant 
image region that
localizes the tumor, which is used to explain the OS time estimation in a post-hoc manner.
Contrary to the pre-hoc pipeline, our method can estimate the OS time and provide the tumor localization without the use of ground truth segmentation labels and of hand-designed radiomic features.

We evaluate our model on the Multimodal Brain Tumor Segmentation Challenge (BraTS) 2019 data set. Results show that our post-hoc model trained with no segmentation labels achieves comparable results to state-of-the-art (SOTA) pre-hoc models that rely on segmentation labels during training.

\section{Related Works}
\label{sec:rel}

OS time prediction is a key task that guides treatment plan for patients suffering from gliomas. With the aim to foster research in OS time prediction, BraTS~\cite{menze_multimodal_2015, bakas_identifying_2019, bakas_advancing_2017} is a publicly available data set that contains the ground truth of OS time and tumor segmentation from brain magnetic resonance images (MRI).

Current SOTA methods for OS time prediction~\cite{agravat2019brain,wang2019automatic,feng2019brain,wang20193d,zhou2020m2net} follow a pre-hoc pipeline consisting of two stages: 1) segmentation of the tumor sub-regions; and 2) use of the tumor segmentation map from 1) to train a OS time prediction model.
In the first stage, the tumor segmentation maps are obtained manually~\cite{zhou2020m2net} or automatically~\cite{agravat2019brain,wang2019automatic,feng2019brain,wang20193d}.
However, both manual and automated approaches require ground truth segmentation labels, which are costly to obtain.
In the second stage, most prediction models use hand-crafted radiomics features~\cite{agravat2019brain,wang2019automatic,feng2019brain,wang20193d}.
These features include shape, intensity and texture features, extracted from the brain image and the tumor sub-regions segmentation.
For the OS time prediction task, \cite{agravat2019brain,wang2019automatic} train a random forest regressor, \cite{feng2019brain} rely on a linear model and \cite{wang20193d} uses a  neural network with two hidden layers.
More recently, Zhou \textit{et al.}~\cite{zhou2020m2net} used image and radiomics features for the OS prediction model, but still relied on segmentation labels to allow the extraction of patches centered at the tumor.

Although the pre-hoc approaches above have advanced the field of OS time prediction from brain MRI images, they require ground truth tumor segmentation labels for training.
Such requirement is a major weakness of these approaches because the segmentations labels are costly to obtain and are, in general, fairly unreliable given that annotations disagreements can be common among experts.
On the contrary, post-hoc methods have been shown~\cite{maicas2019pre} to be competitive with the performance of pre-hoc methods for prediction tasks without requiring  ground truth segmentation for training.
In this paper, we propose a post-hoc approach that performs OS time prediction after being trained using only OS time ground truth.
Additionally, our post-hoc method provides a saliency map that shows the relevant area of the image for the OS prediction, which is expected to be the tumor localization -- such saliency map can be used to explain the OS time prediction.

\section{Method}
\label{sec:method}

\begin{figure}[t]
\centering
\includegraphics[width=1.00\linewidth]{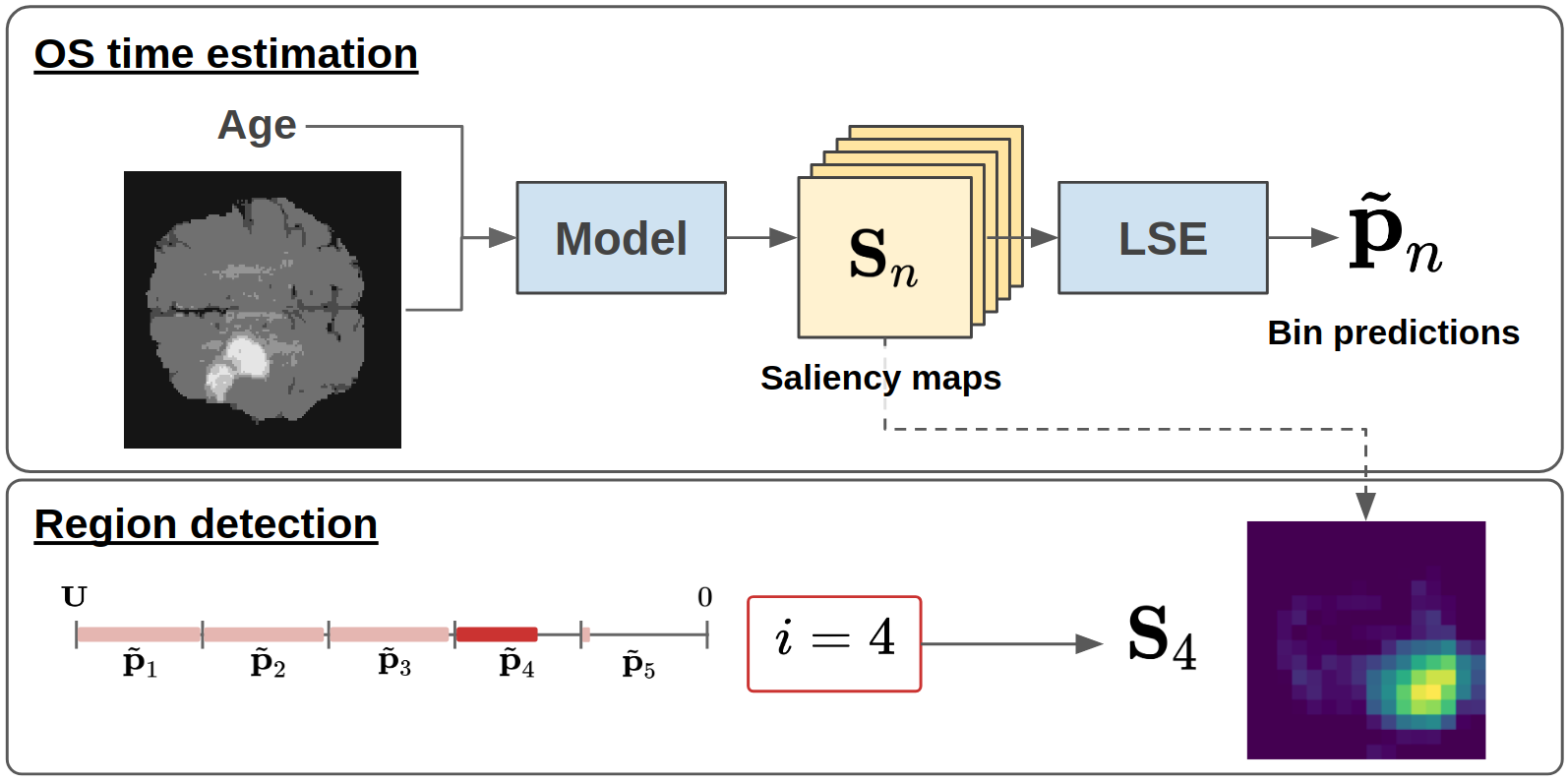}
\caption{
The proposed post-hoc model consists of two parts:
1) OS time estimation, where the model splits the prediction into bins representing an increasing OS time estimation (in days); and
2) region detection, where the saliency maps $\{\mathbf{S}_n\}_{n=1}^{N}$ extracted in 1) are used to highlight relevant regions of the image used for OS time estimation.
The final OS time prediction is obtained from the sum of $\{\tilde{\mathbf{p}}_n\}_{n=1}^N$, with each $\tilde{\mathbf{p}}_n$ denoting the product  of the number of days in bin $n$ and the survival probability -- see Eq.~\ref{eq:pred}.}
\label{fig:model}
\end{figure}

The proposed post-hoc method estimates the number of days of OS time and provides the tumor location from pre-operative brain MRI.
For risk prediction, we split the OS time into $N$ bins, where each bin represents a particular range of days for the estimation.
Note that the OS time estimation will steadily decrease for consecutive bins because if the estimation for 100 days is $X\%$, then the risk for 200 days must be equal or smaller.
The model outputs a survival probability 
associated with each bin, and the final OS time prediction is computed by summing the product between the number of days for each bin and the survival probability
produced by the model.
For model explainability, we use the saliency maps obtained during the process of estimating the risk.
See Fig.~\ref{fig:model} for a graphical summary of our method.
We describe the data set in Sec.~\ref{sec:Data}, and our method in Sec.~\ref{sec:model}.

\subsection{Data Set}
\label{sec:Data}

The data set is defined by
$\mathcal{D} = \{(\mathbf{x}_i,\mathbf{y}_i,\mathbf{s}_i,\mathbf{a}_i)\}_{i=1}^{|\mathcal{D}|}$,
where
$\mathbf{x}_i: \Omega \rightarrow \mathbb{R}^4$ 
denotes a brain MRI image with 4 modalities (Flair, T1, T1ce and T2), and $\Omega$ is the image lattice,
$\mathbf{y}_i \in \mathbb N$ indicates the OS time defined in days, $\mathbf{s}_i: \Omega \rightarrow \mathcal{Y}$, $\mathcal{Y} = \{\textit{Background}, \textit{Tumor}\}$ is a binary map that represents the whole tumor segmentation annotation, and  $\mathbf{a}_i \in \mathbb N$ indicates the age in years.
Note that $\mathbf{s}_i$ is only used for evaluating our proposed weakly supervised tumor segmentation method -- we never use those segmentations for training.

\begin{figure}[t]
\centering
\includegraphics[width=0.90\linewidth]{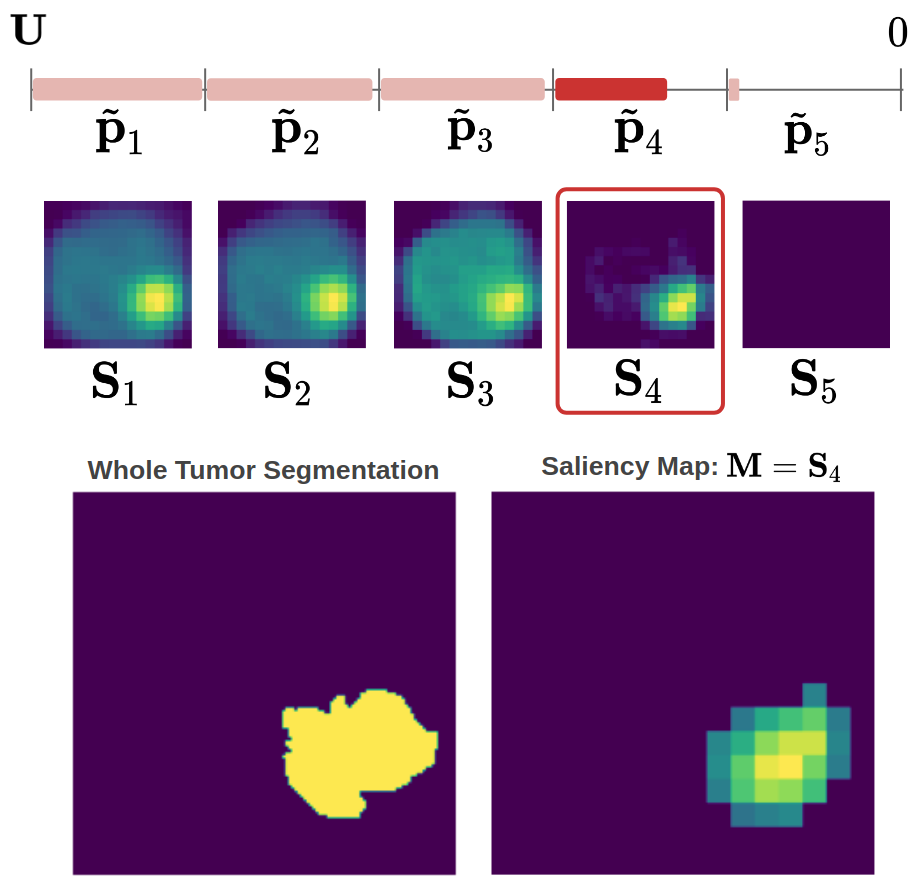}
\caption{
Comparison between the true label segmentation (bottom-left) for the whole tumor and the saliency map extracted from the bin predictions (bottom-right).
The selected saliency map corresponds to the last bin responsible for the prediction (see Eq.~\ref{eq:saliency2}).
}
\label{fig:saliency}
\end{figure}

\subsection{Model}
\label{sec:model}

Our proposed model uses a fully convolutional architecture~\cite{oquab_is_2015} to extract features $\mathbf{F}_i \in \mathbb{R}^{Q \times V^3}$ from $\mathbf{x}_i$, which contains $Q$ feature maps, and a simple linear layer to extract $\tilde{\mathbf{s}}_i \in \mathbb R^Q$ from $\mathbf{a}_i$.
Then $\mathbf{F}_i$ and $\tilde{\mathbf{s}}_i$ are combined as in:
\begin{equation}
    \tilde{\mathbf{F}}_i = \mathbf{F}_i + b(\tilde{\mathbf{s}}_i;V^3),
    \label{eq:features}
\end{equation}
where $b(\tilde{\mathbf{s}}_i;V^3) \in \mathbb{R}^{Q \times V^3}$ takes each $\tilde{\mathbf{s}}_{i,q}$, for $q \in \{1,...,Q\}$, and broadcasts to the respective $V^3$ dimensions.
A final convolution layer takes $\tilde{\mathbf{F}}$ from~\eqref{eq:features} as input to produce $N$ saliency maps $\{ \mathbf{S}_{i,n} \}_{n=1}^{N}$, each of size $V^3$, where $N$ is the number of bins used to discretize $\mathbf{y}_i$.
More specifically, each bin $\mathbf{b}_n$ represents a number of survival days $\{\mathbf{b}_n\}^{N}_{n=1}$ using equidistant values between 0 and the maximum number of survival days in the data set, which is $U=1800$ days.
In fact, the maximum number of survival days represents an upper-limit that is obtained by summing the values for each bin: $U = \sum^{N}_{n=1} {\mathbf{b}_n}$, with $\mathbf{b}_n = U/N$ for all $n\in\{1,...,N\}$.
We apply a global pooling on the saliency maps $\mathbf{S}_{i,n}$ to obtain the predictions for each bin $\{\tilde{\mathbf{p}}_{i,n}\}^{N}_{n=1}$ as follows:
\begin{equation}
    \tilde{\mathbf{p}}_{i,n} = \mathbf{b}_n \cdot \mathbf{p}_{i,n},
    \label{eq:predbin2}
\end{equation}
where
\begin{equation}
    \mathbf{p}_{i,n} = \sigma(LSE(\mathbf{S}_{i,n})),
    \label{eq:predbin1}
\end{equation}
with $\sigma$ denoting the sigmoid activation function, and $LSE$ the log-sum-exp pooling function.

We compute the final OS time prediction $\tilde{\mathbf{y}}_i$ by subtracting the predictions for each bin $\tilde{\mathbf{p}}_{i,n}$ from the upper limit $U$:
\begin{equation}
    \tilde{\mathbf{y}}_i = f(\mathbf{x}_i,\mathbf{a}_i;\theta) = U - \sum\limits^{N}_{n=1} \tilde{\mathbf{p}}_{i,n},
    \label{eq:pred}
\end{equation}
where $f(\mathbf{x}_i,\mathbf{a}_i;\theta)$ represents the whole model parameterized by $\theta$ ($\theta$ denotes all weights of the deep learning model), taking image $\mathbf{x}_i$ and age $\mathbf{a}_i$ as inputs.
The rationale behind (\ref{eq:pred}) is that we expect the saliency maps to be active on the regions responsible for OS time survival. 
Thus a higher $\mathbf{p}_{i,n}$ indicates a lower OS time prediction by the number of days in bin $n$.

\textbf{The training} of our model minimizes the loss function:
\begin{equation}
    \ell_{Total}(\mathcal{D},\theta) = \frac{1}{|\mathcal{D}|} \sum_{i=1}^{|\mathcal{D}|} \ell_{mae}(\tilde{\mathbf{y}}_i,\mathbf{y}_i) + \alpha \ell_{pen}(\mathbf{p}_i),
    \label{eq:finalLoss}
\end{equation}
where $\ell_{mae}(\tilde{\mathbf{y}}_i,\mathbf{y}_i)$ is the mean absolute error (MAE) between the ground truth label $\mathbf{y}_i$ and the model prediction $\tilde{\mathbf{y}}_i$ for OS time in~\eqref{eq:pred}, $\ell_{pen}(\mathbf{p}_i)$ is a smoothness term that penalizes inconsistent predictions between neighboring bins which 
forces consecutive bins of the $i$-th training prediction $\mathbf{p}_i$ (defined in Eq.~\ref{eq:predbin2}) to represent a risk estimation in descending order, with minimal amount of transition between neighboring bins.
In particular, in this loss, a bin $n$ of $\mathbf{p}_i$ should be active only when all previous bins from $1$ to $n-1$ have become active.
Therefore, the penalization term to force this constrain is defined as follows:
\begin{equation}
    \ell_{pen}(\mathbf{p}_i) = \frac{1}{N-1} \sum\limits^{N-1}_{n=1} \max(0, (\mathbf{p}_{i,n+1} - \mathbf{p}_{i,n})),
    \label{eq:l2}
\end{equation}
where $\mathbf{p}_{i,n}$ represents the prediction of the $i$-th training sample at bin $n$.
The weight $\alpha$ in~\eqref{eq:finalLoss} controls the smoothness strength.

For the explanation of the predictions of the post-hoc model, we observe that the saliency maps $\{\mathbf{S}_{i,n}\}_{n=1}^N$ should contain the interpretation of the OS time prediction, represented by the tumor localization.
We use the saliency map $\mathbf{S}_{i,n^*}$, where $n^*$ corresponds to the index of the last bin responsible for the prediction from the model (see Fig.~\ref{fig:saliency}):
\begin{equation}
    n^* = \underset{n \in \{1,...,N\}}{\operatorname{argmax}} \{\min(\mathbf{p}_{i,n}, |1 - \mathbf{p}_{i,n}))\}
    \label{eq:saliency2}
\end{equation}

\textbf{The inference} of our method uses the forward pass of the trained model and gives the final OS time prediction using~\eqref{eq:pred} and the saliency map from~\eqref{eq:saliency2}.

\section{Experiments}
\label{sec:exp}

\begin{figure}[t]
\centering
\includegraphics[width=1.00\linewidth]{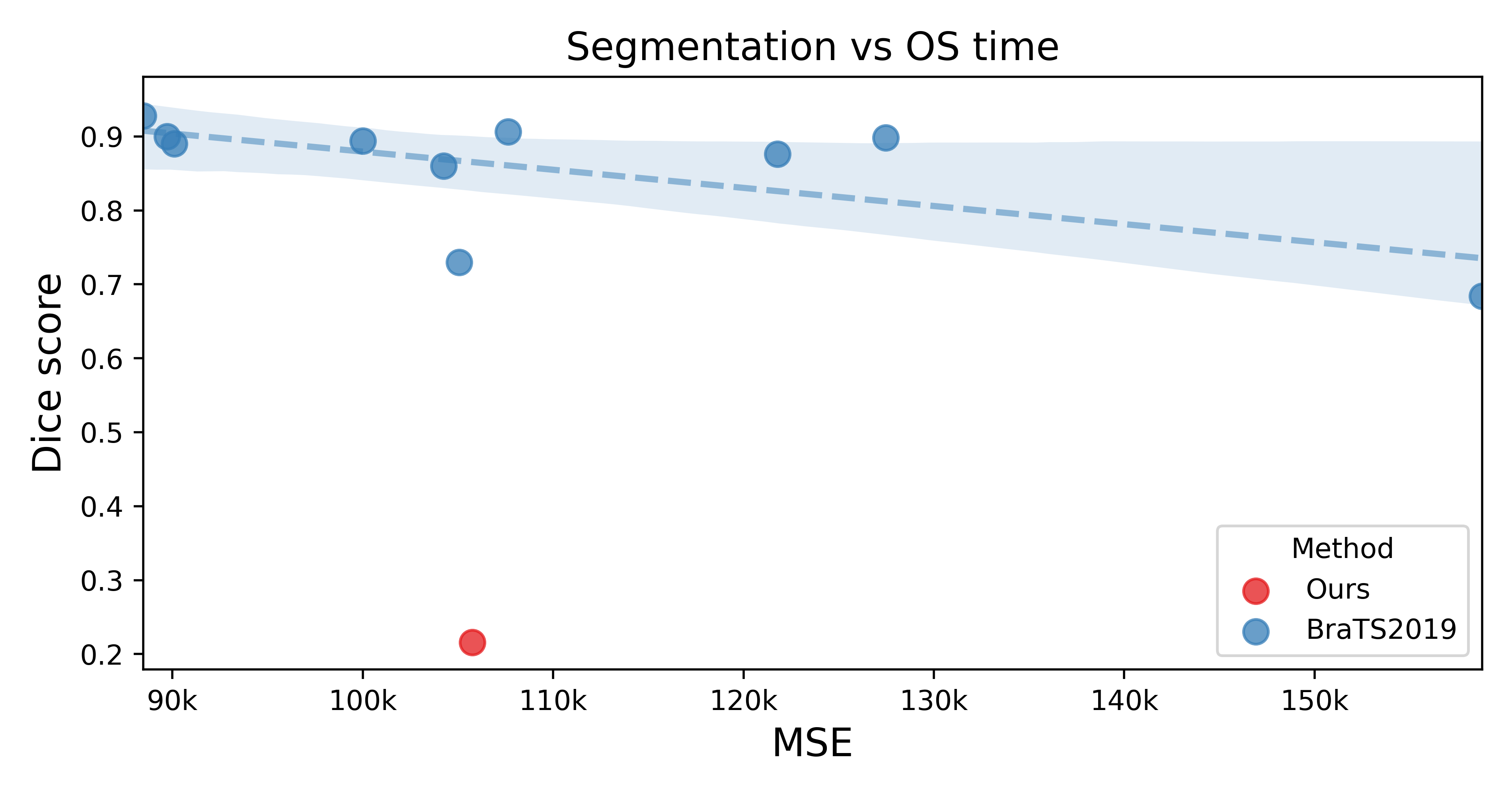}
\caption{
Comparison of survival (MSE - smaller is better) and segmentation results (Dice - larger is better)  on BraTS 2019 validation set~\cite{crimi2020brainlesion}.
Note the weak correlation between segmentation and OS results and that our method (red dot) has a competitive MSE even with a low Dice result.
}
\label{fig:comp}
\end{figure}

\subsection{Data Set}

We conduct the experiments in this paper using the data set from the Multimodal Brain Tumor Segmentation Challenge (BraTS) 2019 \cite{menze_multimodal_2015, bakas_advancing_2017, bakas_identifying_2019}.
The data set contains pre-operative MRI scans of glioblastoma (GBM/HGG) and low-grade glioma (LGG) and focuses on the segmentation of brain tumors and the prediction of patient OS time.
As the focus of this method is OS time prediction, we use a subset of 210 cases that includes survival labels for training, and we use 20\% as internal validation for model selection (i.e., hyper-parameter estimation).
Also, a validation set of 29 cases is available for the OS time prediction task, while ground truth annotations are not available for this set, it is possible to evaluate the results using a provided web service for the challenge.
Each case includes MRI images for four  modalities: Flair, T1, T1ce and T2.
The ground truth for segmentation is marked with three labels:
necrotic and non-enhancing tumor (label 1),
edema tumor (label 2),
and enhancing tumor (label 4).
For evaluation purposes we use the complete tumor extent referred to as "whole tumor", as the saliency map we obtain does not distinguish between labels.
For the task of OS prediction, each case includes the age of patient, a resection status: Gross Total Resection (GTR) or Subtotal Resection (STR), and the OS defined in days.

\subsection{Experimental Set Up}

Each modality of the MRI image and the age are normalized with mean value of zero and standard deviation of one, then, the image data of the four modalities for each case is down-sampled and stacked into a multi-channel data with size $128^3 \times 4$.
During training we apply random scaling between 1 and 1.1 as data augmentation.
The model consists of 4 blocks of: convolution, activation and batch normalization, where
the activation function is a Leaky ReLU~\cite{maas_rectifier_2013} with a 0.1 negative slope.
The model is trained using the Adam~\cite{kingma_adam:_2015} optimizer with a momentum of 0.9, weight decay of 0.001, a mini-batch size of 8 and $\alpha$ of 10,000.

We use the Mean Squared Error (MSE) and classification accuracy to evaluate our model. For the classification, we use the official evaluation setup~\cite{bakas_identifying_2019} that splits the OS time into three groups: 1) short-survivors (less than 10 months), 2) mid-survivors (between 10 and 15 months), and 3) and long-survivors (more than 15 months). To assess the interpretability of our saliency maps, we evaluate the weakly supervised segmentation by using a dice score between the top 5\% activations of the saliency map $\mathbf{S}_{i,n^*}$ from~\eqref{eq:saliency2} and the ground truth segmentation for the whole tumor.

\subsection{Results}

In Table~\ref{tab:res1} we compare the OS time predictions of our method and the best four results from the BraTS 2019 competition~\cite{agravat2019brain,wang2019automatic,feng2019brain,wang20193d}.
Note that our proposed post-hoc method only uses OS time labels during the training process, while the rest of methods follow a pre-hoc pipeline that requires segmentation labels and hand-designed radiomics features.
We also compare segmentation versus OS time prediction results published on BraTS 2019~\cite{crimi2020brainlesion} in Fig.~\ref{fig:comp} and note that better segmentation accuracy does not imply better OS time prediction accuracy.
This is an important point because even though the segmentation result produced by our weakly-supervised saliency map (with dice around 0.2) is much worse than the fully supervised SOTA results (dice $>$ 0.9), our OS time prediction is competitive with the SOTA.
In Tab.~\ref{tab:abla} we provide an ablation study to examine the effects of including the age of the patient as an input to the model.
Moreover, in Tab.~\ref{tab:abla} we also study an alternative method for our proposed survival function, based on a simple regressor with a similar model architecture.

\begin{table}[]
\centering
\caption{
Comparison on OS time prediction results on BraTS 2019 validation set.
Note that our method uses only survival labels while the others use segmentation and survival labels for training.
The evaluation metrics are: accuracy, mean squared error, median squared error, standard squared error and SpearmanR.
}
\scalebox{0.894}{
\begin{tabular}{c|c|c|c|c|c}
\hline
Paper & Acc & MSE & medianSE & stdSE & SpearmanR \\
\hline
\cite{agravat2019brain} & 0.586 & 105,062 & 16,461 & 188,752 & 0.404 \\
\cite{wang2019automatic} & 0.590 & ~~89,724 & 36,121 & - & 0.360 \\
\cite{feng2019brain} & 0.310 & 107,639 & 77,906 & 109,587 & 0.204 \\
\cite{wang20193d} & 0.448 & 100,000 & 49,300 & 135,000 & 0.250 \\
Ours & 0.517 & 105,746 & 51,962 & 181,311 & 0.248 \\
\hline
\end{tabular}
}
\label{tab:res1}
\end{table}

\begin{table}[]
\centering
\caption{
Ablation experiments on the use of age for OS time prediction and a simple regression method vs our proposed post-hoc method. Acc refers to accuracy and MSE to mean squared error.
}
\scalebox{0.894}{
\begin{tabular}{c|c|c|c|c}
\hline
\multirow{2}{*}{Method} & \multicolumn{2}{c|}{Train set} & \multicolumn{2}{c}{Validation set} \\
\cline{2-5}
       & Acc & MSE & Acc & MSE \\
\hline
Regression & 0.286 &  73,878 & 0.207 &  223,604 \\
\hline
Regression + Age & 0.405 &  75,241 & 0.379 & 140,457 \\
\hline
Post-hoc & 0.405 &  77,439 & 0.379 &  134,213 \\
\hline
Post-hoc + Age & \textbf{0.548} &  \textbf{57,198} & \textbf{0.517} & \textbf{105,746} \\
\hline
\end{tabular}
}
\label{tab:abla}
\end{table}

\section{Discussion and Conclusion}
\label{sec:disc}

In this paper, we proposed the use of a post-hoc method for OS time prediction from brain MRI and patient demographics (represented by age) of cases with high-grade gliomas.
SOTA methods for OS time prediction follow a pre-hoc pipeline that requires for training the OS time labels and tumor segmentation annotations which are costly to obtain and may contain annotation inconsistencies.
On the contrary, as shown in Table~\ref{tab:res1}, our proposed post-hoc method is competitive to SOTA pre-hoc methods and only requires the OS time labels for training -- in addition, it produces a saliency map localizing the tumor in the image for explaining the OS time estimation.
Our post-hoc method also outperforms a regression model using a similar architecture, as shown in Table~\ref{tab:abla}.
Note that as shown in Fig.~\ref{fig:comp}, extracting features using better segmentation results does not imply an increase on OS time prediction accuracy.
We leave for future work the design of a post-hoc method that can improve both the OS time prediction accuracy as well as the precision of the saliency map segmentation.

\section*{Compliance with Ethical Standards } 

This research study was conducted retrospectively using human subject data made available in open access by~\cite{bakas_segmentation_2017,bakas_segmentation_2017-1,clark_cancer_2013}. Ethical approval was not required as confirmed by the license attached with the open access data. 

\section*{Acknowledgments}

\thanks{This work was supported by the Australian Research Council through grants DP180103232 and FT190100525.}

\bibliographystyle{IEEEbib}
\bibliography{zotero.bib}

\end{document}